\documentclass[a4paper,conference]{IEEEtran}
\IEEEoverridecommandlockouts
\ifCLASSINFOpdf
\usepackage{graphicx}
\usepackage{comment}
\usepackage{amsmath,amssymb} 
\usepackage{subfigure}
\usepackage{multirow}
\usepackage{verbatim}
\usepackage{color}
\usepackage{xspace}
\usepackage[colorlinks,urlcolor=blue]{hyperref}

\DeclareRobustCommand*{\IEEEauthorrefmark}[1]{%
	\raisebox{0pt}[0pt][0pt]{\textsuperscript{\footnotesize\ensuremath{#1}}}}

\makeatletter
\DeclareRobustCommand\onedot{\futurelet\@let@token\@onedot}
\def\@onedot{\ifx\@let@token.\else.\null\fi\xspace}

\def\eg{\emph{e.g}\onedot} 
\def\ie{\emph{i.e}\onedot} 
 
\def\etc{\emph{etc}\onedot} 
 
\def\etal{\emph{et al}\onedot}
\makeatother

\else
  
\fi

\begin{document}
%
\title{Adaptive Dilated Convolution For Human Pose Estimation}

\author{
	\IEEEauthorblockN{
		Zhengxiong Luo$^{1,2,3,4,5}$,
		Zhicheng Wang$^{1}$,
		$^*$Yan Huang$^{3,4,5}$, 
		Liang Wang$^{3,4,5}$, 
		Tieniu Tan$^{3,4,5}$,
		Erjin Zhou$^{1}$,\\
	}
	\IEEEauthorblockA{
		\IEEEauthorrefmark{1} Megvii Inc \quad
		\IEEEauthorrefmark{2} University of Chinese Academy of Sciences (UCAS) \\
		\IEEEauthorrefmark{3} Center for Research on Intelligent Perception and Computing (CRIPAC)\\
		\IEEEauthorrefmark{4} National Laboratory of Pattern Recognition (NLPR) \quad
		\IEEEauthorrefmark{5} Institute of Automation, Chinese Academy of Sciences (CASIA)\\
		zhengxiong.luo@cirpac.ia.ac.cn \quad \{wangzhicheng, zej\}@megvii.com \quad \{yhuang, wangliang, tnt\}@nlpr.ia.ac.cn
	}
}

\maketitle
\renewcommand{\thefootnote}{\fnsymbol{footnote}}
\footnotetext[1]{Corresponding author}

	\begin{abstract}
	
	Most existing human pose estimation (HPE) methods exploit multi-scale information by fusing feature maps of four different spatial sizes, \ie $1/4$, $1/8$, $1/16$, and $1/32$ of the input image. There are two drawbacks of this strategy: 1) feature maps of different spatial sizes may be not well aligned spatially, which potentially hurts the accuracy of keypoint location; 2) these scales are fixed and inflexible, which may restrict the generalization ability over various human sizes. Towards these issues, we propose an adaptive dilated convolution (ADC). It can generate and fuse multi-scale features of the same spatial sizes by setting different dilation rates for different channels. More importantly, these dilation rates are generated by a regression module. It enables ADC to adaptively adjust the fused scales and thus ADC may generalize better to various human sizes. ADC can be end-to-end trained and easily plugged into existing methods. Extensive experiments show that ADC can bring consistent improvements to various HPE methods. The source codes will be released for further research. 
	
\end{abstract}

\section{Introduction}

Human Pose Estimation (HPE) aims to locate skeletal keypoints (\eg ear, shoulder, elbow, \etc) of all persons in the given RGB image. It is fundamental to action recognition and has wide applications in human-computer interaction, animation, \etc. This paper is interested in single-person pose estimation, which is the basis of multi-person pose estimation~\cite{stacked,cpm}.

HPE involves two sub-tasks: location (determining where the keypoints are) and classification (determining which kinds the keypoints are). The location needs plenty of local details to get pixel-level accuracy. While classification requires a relatively larger receptive field to extract discriminative semantic representations~\cite{rsn}. Consequently, HPE methods have to fuse multi-scale information to make a balance between these two sub-tasks~\cite{hrnet}. Most nowadays HPE methods~\cite{cpn,scarb,mspn,danet,swahr} repeatedly downscale feature maps to enlarge the receptive fields. Feature maps of different spatial sizes (\ie $1/4$, $1/8$, $1/16$, and $1/32$ of  the input image) are then resized and summed to exploit multi-scale information. 

\begin{figure}
	\centering
	\includegraphics[width=\linewidth]{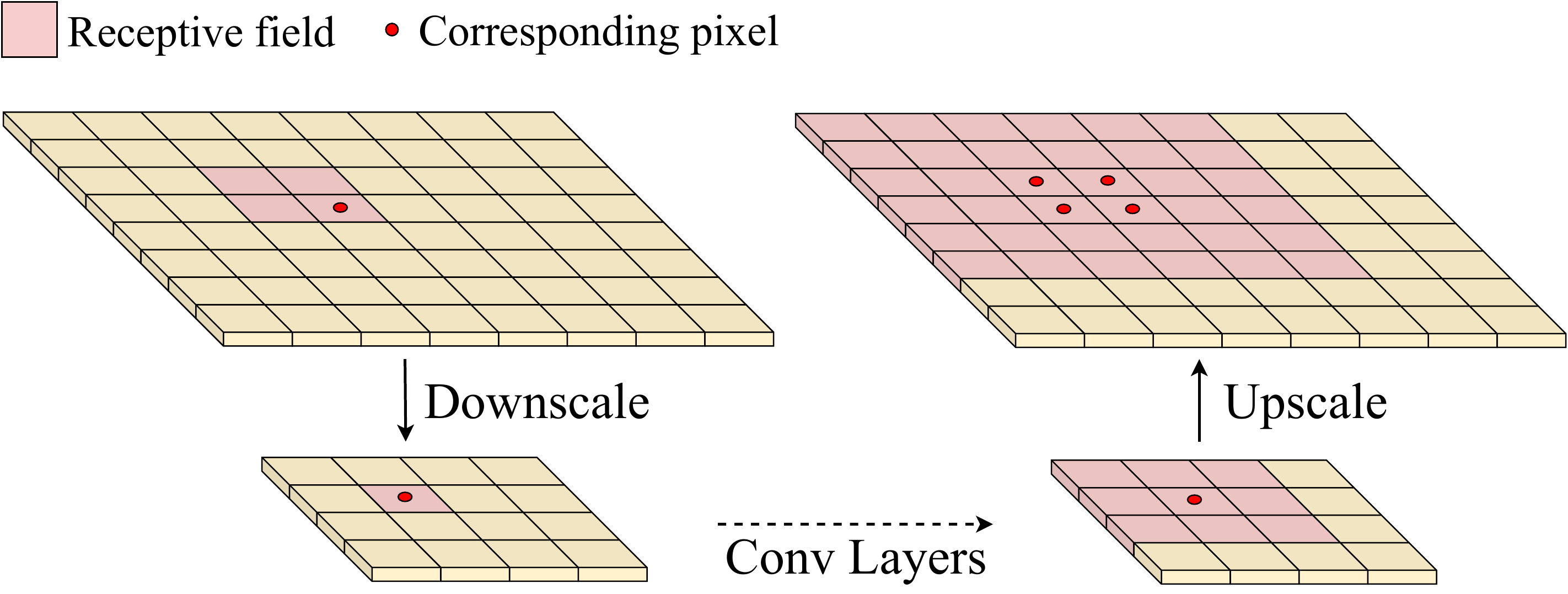}
	\caption{The receptive fields can be easily enlarged by the \textit{downscale-conv-upscale} loop. But during the upscaling, there will multiple possible postilions for the corresponding pixel. The upscaled feature maps may be not well aligned with original ones.}\label{align}
\end{figure}

This strategy has made great achievements in HPE~\cite{mspn,hrnet,danet}, but it still leaves to be desired. In this strategy, feature maps are downscaled by strided convolution (or pooling). As shown in Figure~\ref{align}, during the downscaling, multiple pixels on the larger feature maps are merged into the same pixel on the smaller ones. The location information will be destroyed in this process. While during the upscaling, even if the transposed convolution~\cite{deconv} is used, it is hard to recover the destroyed location information. Consequently, there will be multiple possible corresponding positions on the upscaled feature maps for original single pixel. Although the final resized feature maps have the same spatial sizes, their pixels may be not well aligned. This spatial non-alignment potentially hurts the accuracy of location. Thus, it may be more preferred to fuse multi-scale features of the same spatial sizes.

An alternative method is to use dilated convolution, instead of downscaling, to enlarge receptive fields. In~\cite{multi_dialation,aspp}, multiple convolutional layers with different dilation rates are used to extract feature maps at different scales. These feature maps have the same spatial sizes and are well aligned spatially. They are concatenated and fused by $1\times1$ convolution to exploit multi-scale information. However, these dilation rates are still manually set and fixed, which may restrict the generalization ability over various human sizes.

Towards these issues, we propose an adaptive dilated convolution (ADC) in this paper. As shown in Figure~\ref{adc}, it divides channels into different dilation groups and uses a dilation-rates regression module to adaptively generate dilation rates for these groups. Compared with previous multi-scale fusion methods, ADC has three advantages: \textit{i}) Instead of using multiple independent dilated convolution layers, ADC directly assigns different dilation rates to its channels. In this way, ADC can generate and fuse multi-scale features in a single layer, which is more elegant and efficient. \textit{ii}) ADC allows fractional dilation rates, which enables ADC to adjust receptive fields with finer granularity, instead of only four fixed integer scales. Thus ADC may be able to exploit richer and finer multi-scale information. \textit{iii}) The dilation rates in ADC are adaptively generated, which could help ADC to generalize better to various human sizes. 

ADC can be easily plugged into existing HPE methods and trained end-to-end by standard back-propagation. Our contributions can be summarized into three points:
\begin{itemize}
	\item[1.] We attempt to address the spatial non-alignment and inflexibility problems in nowadays multi-scale fusion methods of  HPE. These problems are important to the accuracy of  location and generalization ability over various human sizes.
	\item[2.] We propose an adaptive dilated convolution (ADC), which could flexibly fuse well-aligned multi-scale features in a single convolutional layer by adaptively generating dilation rates for different channels.
	\item[3.] The proposed ADC can be easily plugged into existing HPE methods and extensive experiments show that ADC can bring these methods consistent improvements.
\end{itemize}

\section{Related Works}
\subsection{Multi-scale Fusion}

Multi-scale fusion is widely adopted in many high-level vision tasks, such as detection~\cite{fpn,deepfpn,nasfpn}, semantic segmentation~\cite{panet}, \etc. On the one hand, these tasks involve both location and classification. They need multi-scale information to make a balance between these two sub-tasks. On the other hand, these tasks need to tackle objects of various sizes. They scale-invariant representations to get more stable performances. In these tasks, most methods~\cite{mspn,fpn,panet} firstly extract a feature pyramid, which contains feature maps of different spatial sizes, and then fuse feature maps to obtain multi-scale information. However, as we have discussed above, the fused features may be not well spatially-aligned. This non-alignment may hurt the accuracy of location. For detection and segmentation, this influence could be ignored, because their evaluation metrics (IOU) are less sensitive to the accuracy of location. While HPE methods are evaluated by OKS, which will be heavily influenced by pixel-level errors. Thus the non-alignment may restrict the performance of HPE methods. In the proposed adaptive dilated convolution, multi-scale features are of the same spatial sizes, which may be more friendly to human pose estimation.

\subsection{Dilated Convolution}

The main idea of dilated convolution is to insert zeros between pixels of convolution kernels. It is widely used in segmentation~\cite{deeplabv3+,deeplab} to enlarge the receptive fields while keeping the resolutions of feature maps. As the size of its receptive field can be easily changed by adjusting its dilation rate, dilated convolution is also used to aggregate multi-scale context information. For example, in~\cite{multi_dialation}, the outputs of convolutional layers with different dilation rates are fused to exploit multi-scale context information. And in~\cite{aspp}, a similar idea is adopted in an atrous spatial pyramid pooling (ASPP) module. However, these dilation rates of different layers are manually set and can only be integers, which are not flexible enough. Instead, the dilation rates in ADC can be fractional and are adaptively generated, which enables it to learn more suitable receptive fields for objects of various sizes. Besides, every dilation group in ADC can represent features at a scale, which enables ADC to fuse richer multi-scale information yet in a simpler way.

\begin{figure}[t]
	\centering
	\includegraphics[width=\linewidth]{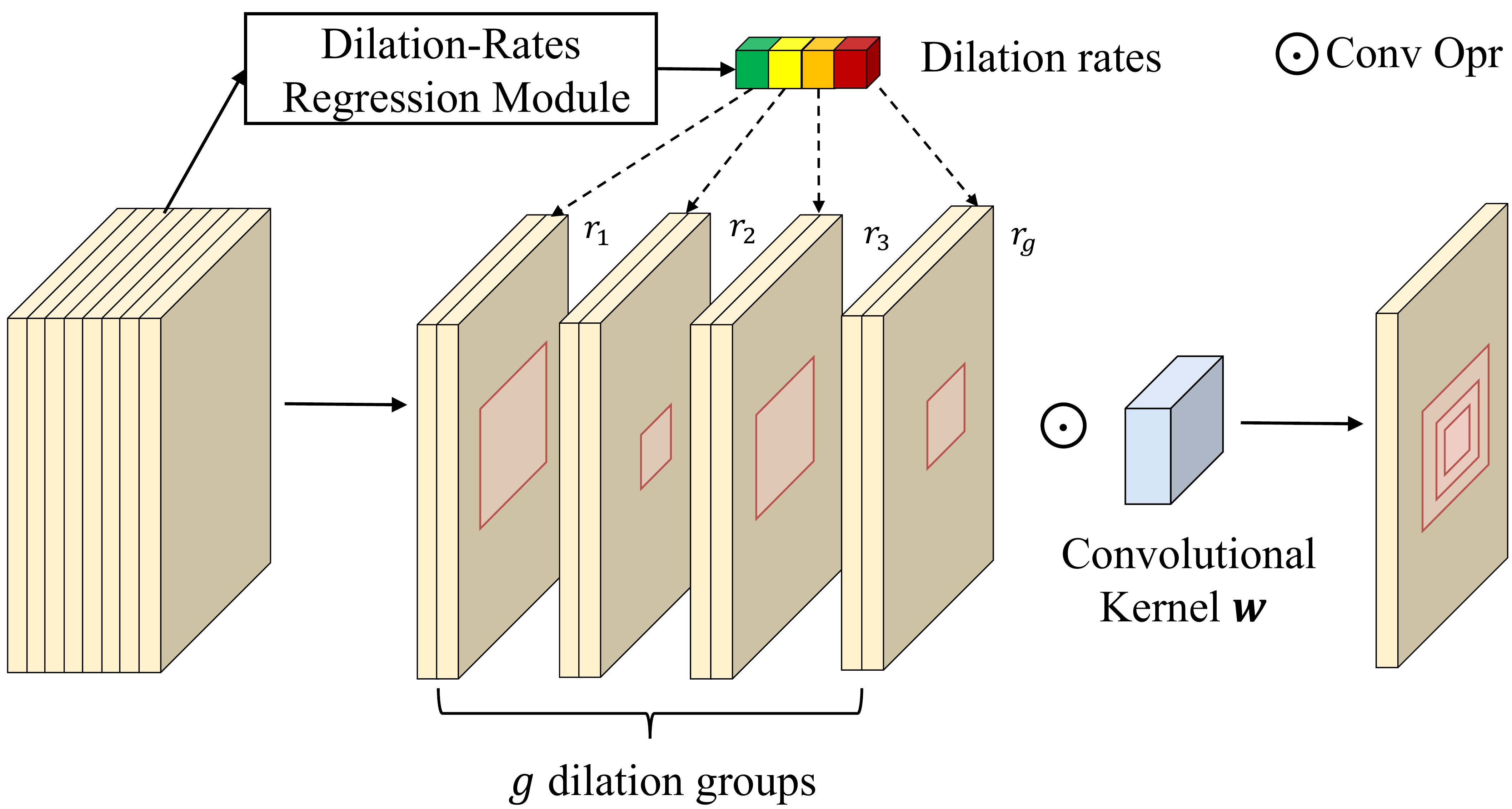}
	\caption{Details of adaptive dilated convolution. Different dilation groups have different dilation rates, and thus have different receptive fields.}\label{adc}
\end{figure}

\section{Adaptive Dilated Convolution}
\subsection{Constant Dilation Rates}
\begin{figure}[t]
	\centering
	\includegraphics[width=\linewidth]{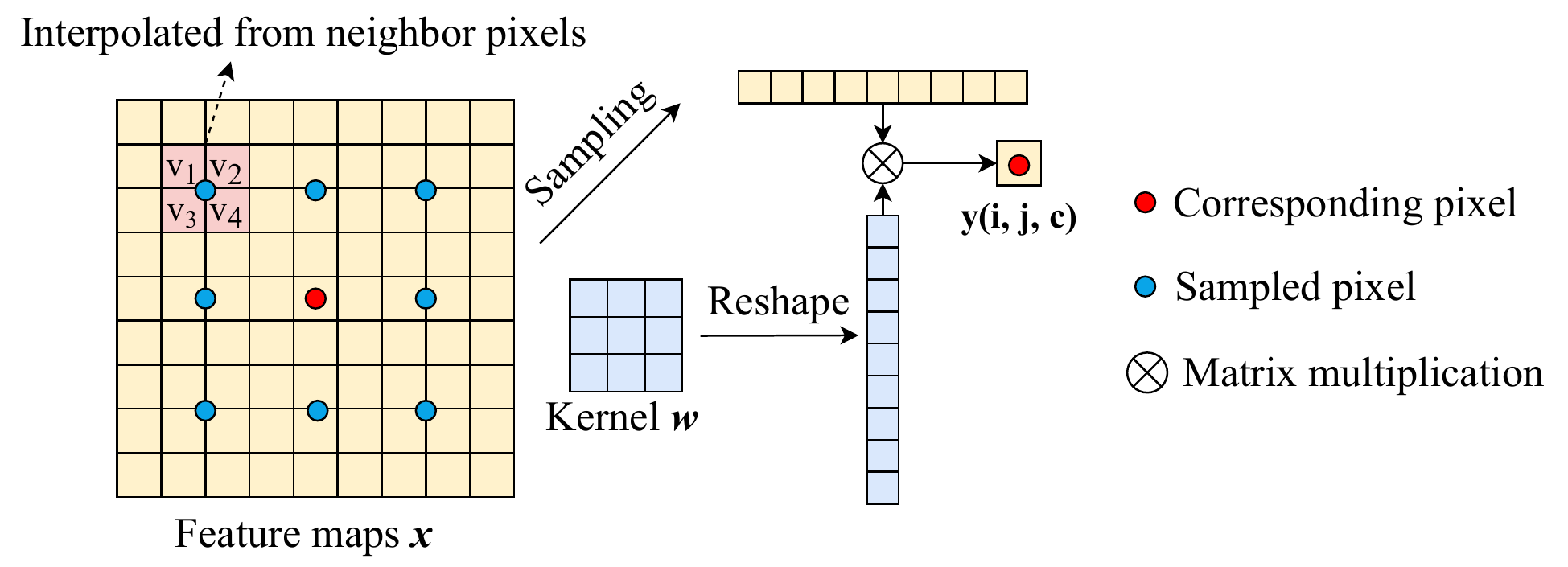}
	\caption{Original dilated convolution can be decomposed into two steps: sampling and matrix multiplication. But in ADC, the dilation rate could be fractional ($2.5$ in the figure), in which case, the sampled values will be interpolated from their neighbor pixels.}\label{dilaed_conv}
\end{figure}

As shown in Figure~\ref{dilaed_conv}, original dilated convolution can be decomposed into two steps: 1) sampling according to a index set $\mathcal{I}$ over the input feature map $\mathbf{x}$; 2) matrix multiplication of the sampled values and convolutional kernel $\mathbf{w}$. The index set $\mathcal{I}$ is defined by the dilation rate $r$ and size of kernel $k\times k$:
\begin{equation}
	\begin{gathered}
		\mathcal{I} =\{ (i\cdot r, j\cdot r, c) \}, \\
		s.t. \quad \lfloor -k/2 \rfloor \leq i, j\leq \lfloor k/2 \rfloor, \quad 0 \leq c < C_{in},
	\end{gathered}
\end{equation}
where $C_{in}$ is the number of channels in $x$, $\lfloor \cdot \rfloor$ denotes rounding down to the nearest integer. Specially, if $k=3$ and $C_{in}=1$ then
\begin{equation}
	\begin{aligned}
		\mathcal{I}=&\{(-r,-r, 0), (-r, -r+1, 0),\dots,\\
		&(r, r-1, 0), (r, r, 0)\}.
	\end{aligned}
\end{equation}
For value at location $(i, j, c)$ of the output feature map $\mathbf{y}$, we have
\begin{equation}
	\mathbf{y}(i, j, c) = \sum_{\Delta\mathbf{p}\in \mathcal{I}} \mathbf{w}^c(\Delta\mathbf{p}) \cdot \mathbf{x}((i, j, 0) + \Delta\mathbf{p}),
\end{equation}
where $\Delta\mathbf{p}$ enumerates the indexes in $\mathcal{I}$, and $\mathbf{w}^c$ denotes the corresponding convolutional kernel for the $c^{th}$ output channel.

The receptive field for each channel in convolutional layer is defined as the square covered by index set $\mathcal{I}$. In original dilated convolution, the receptive fields of all channels are the same. Their sizes are:
\begin{equation}
	\begin{aligned}
		Area=&(\lfloor k/2\rfloor \cdot r - \lfloor -k/2 \rfloor\cdot r)^2\\
		=&(kr - r +1)^2.
	\end{aligned}
\end{equation}
Specially, when $r=1$, the size of receptive field is $k^2$.


\subsection{Adaptive Dilation Rates}
In adaptive dilated convolution, the dilation rates are no longer manually set. As shown in Figure~\ref{adc}, we use a dilation-rates regression module (DRM) to adaptively generate the dilation rates for different channels. DRM consists of a global average pooling layer and two fully connected layers with nonlinear activations. Suppose DRM is denoted as a function $\phi(\cdot)$, then generated dilation rate $\mathbf{r}$ is
\begin{equation}
	\mathbf{r} = \phi(\mathbf{x}).
\end{equation}

We divide the input channels into $g$ \textit{dilation groups}. Each group contains $C_{in} / g$ channels. The channels in the same group shares the same dilation rate. Thus the shape of $\mathbf{r}$ is $g \times 1$. And the dilation rate of the $c^{th}$ input channel is $\mathbf{r}^{\lfloor c/g \rfloor}$. If $g=C_{in}$, then each channel has its own dilation rate. If $g=1$, then all channels share the same dilation rate. 

Consequently, the sampling index set becomes
\begin{equation}
	\begin{gathered}
		\mathcal{I} =\{ (i\cdot \mathbf{r}^{\lfloor c/g \rfloor}, j\cdot \mathbf{r}^{\lfloor c/g \rfloor}, c) \} \\
		s.t. \quad -k/2 \leq i, j\leq k/2, \quad 0 \leq c < C_{in}.
	\end{gathered}
\end{equation}

In cases where $\mathbf{r}$ is fractional, as shown in Figure~\ref{dilaed_conv}, we use bilinear interpolation to get the sampling values. Suppose $M(\mathbf{x}, (i, j, c))$ denotes the interpolated value at $(i, j, c)$ on $\mathbf{x}$, then we have:
\begin{equation}
	\mathbf{y}(i, j, c) = \sum_{\Delta\mathbf{p}\in \mathcal{I}} \mathbf{w}^c(\Delta\mathbf{p}) \cdot M(\mathbf{x}, (i, j, 0) + \Delta\mathbf{p}).
\end{equation}

Similarly, in the $c^{th}$ channel of adaptive dilated convolution, the size of receptive field is:
\begin{equation}
	\begin{aligned}
		Area=&(\lfloor k/2\rfloor \cdot \mathbf{r}^{\lfloor c/g \rfloor} - \lfloor -k/2\rfloor \cdot \mathbf{r}^{\lfloor c/g \rfloor})^2\\
		=&(k\cdot \mathbf{r}^{\lfloor c/g \rfloor}- \mathbf{r}^{\lfloor c/g \rfloor} +1)^2.
	\end{aligned}
\end{equation}
Consequently, different dilation groups have different sizes of receptive fields. And thus ADC can fuse multi-scale information in a single layer.

Since all involved operators are numerical differentiable~\cite{dcn,scn}, the proposed adaptive dilated convolution can be easily plugged into existing models trained end-to-end by standard back-propagation.

\subsection{Analysis and Discussion}\label{dis_deform}
\noindent
\textbf{Comparison with Yu \etal}
In~\cite{multi_dialation}, Yu \etal use multiple dilated convolutional layers with different dilation rates to extract features at different scales. ADC adopts a similar idea, but implements it in a simple yet efficient way. Firstly, ADC consists of only one convolutional layer. It does not use independent dilated convolutional layers or extra concatenation. Thus ADC is much more computation-economic and time-saving. Secondly, every dilation group in ADC represents features at a different scale, which enables ADC to exploit richer multi-scale information than~\cite{multi_dialation}. Thirdly, the dilation rates in ADC can be fractional and are adaptively generated, instead of manually set integers. It helps ADC to generalize better to persons of various sizes.

\vspace{0.02\linewidth}\noindent
\textbf{Comparison with deformable convolution.}
In~\cite{dcn}, Dai \etal propose a deformable convolutional layer, which allows the sampling index set $\mathcal{I}$ to be non-grid and irregular. It assigns an offset for each index in  $\mathcal{I}$, instead of only modifying the dilation rates. Compared with adaptive dilated convolution,  deformable convolution enjoys higher degrees of freedom, but it also has a much higher computational cost. More importantly, the offsets introduced in deformable convolution are completely unconstrained and independent. This may cause the input and output feature maps to lose their spatial correspondence, which also potentially hurts the accuracy of location. We also experimentally proved in Sec~\ref{exp_deforms} that the proposed adaptive dilated convolution is more suitable to human pose estimation than deformable convolution.

\vspace{0.02\linewidth}\noindent
\textbf{Comparison with scale-adaptive convolution.}
In~\cite{sac}, Zhang \etal propose a scale-adaptive convolution (SAC) to address inconsistent predictions of large objects and invisibility of small objects in scene parsing. SAC adaptively generates pixel-wise dilation rates to acquire flexible-size receptive fields along spatial dimensions. It works well for scene parsing, which needs to tackle objects of various sizes within a single image. However, in single person pose estimation, size inconsistent across different images plays a more important role, which could be better alleviated via multi-scale fusion along channel dimension. In SAC, different pixels can have different sizes of receptive fields, but different channels share the same dilation rates. Consequently, ADC may be more suitable for single person pose estimation than SAC. In Sec~\ref{exp_deforms}, we also experimentally prove that ADC works better than SAC in HPE methods.

\begin{table*}[t]
	\centering
	\caption{Results of different models with or without adaptive convolution. The input sizes are $256\times192$. Results are reported on COCO val2017.}~\label{aba_adc}
	\setlength{\tabcolsep}{0.7cm}
	\resizebox{\linewidth}{!}{
		\begin{tabular}{c|c|c|ccccc}
			\hline
			Method                                                                     & Backbone                   & ADC      & $AP$   & $AP^{50}$ & $AP^{75}$ & $AP^{M}$ & $AP^{L}$ \\ \hline
			\multirow{6}{*}{\begin{tabular}[c]{@{}c@{}}Simple Baseline\\ \cite{simplebaseline}\end{tabular}} & \multirow{2}{*}{Res50}     & $\times$ & $70.4$ & $88.6$    & $78.3$    & $67.1$   & $77.2$   \\
			&                            & $\surd$  & $71.8$ & $89.2$    & $79.7$    & $68.5$   & $78.7$   \\ \cline{2-8} 
			& \multirow{2}{*}{Res101}    & $\times$ & $71.4$ & $89.3$    & $79.3$    & $68.1$   & $78.1$   \\
			&                            & $\surd$  & $72.5$ & $89.5$    & $80.4$    & $69.3$   & $79.3$   \\ \cline{2-8} 
			& \multirow{2}{*}{Res152}    & $\times$ & $72.0$ & $89.3$    & $79.8$    & $68.7$   & $78.9$   \\
			&                            & $\surd$  & $72.8$ & $89.3$    & $80.6$    & $69.5$   & $79.7$   \\ \hline
			\multirow{4}{*}{\begin{tabular}[c]{@{}c@{}c}HRNet\\ \cite{hrnet}\end{tabular}}                                                     & \multirow{2}{*}{HRNet-W32} & $\times$ & $74.4$ & $90.5$    & $81.9$    & $70.8$   & $81.0$   \\
			&                            & $\surd$  & $75.0$ & $90.6$    & $82.0$    & $71.4$   & $81.7$   \\ \cline{2-8} 
			& \multirow{2}{*}{HRNet-W48} & $\times$ & $75.1$ & $90.6$    & $82.2$    & $71.5$   & $81.8$   \\
			&                            & $\surd$  & $75.5$ & $90.8$    & $82.3$    & $72.3$   & $82.5$   \\ \hline
	\end{tabular}}
\end{table*}
\begin{figure}
	\centering
	\includegraphics[width=\linewidth]{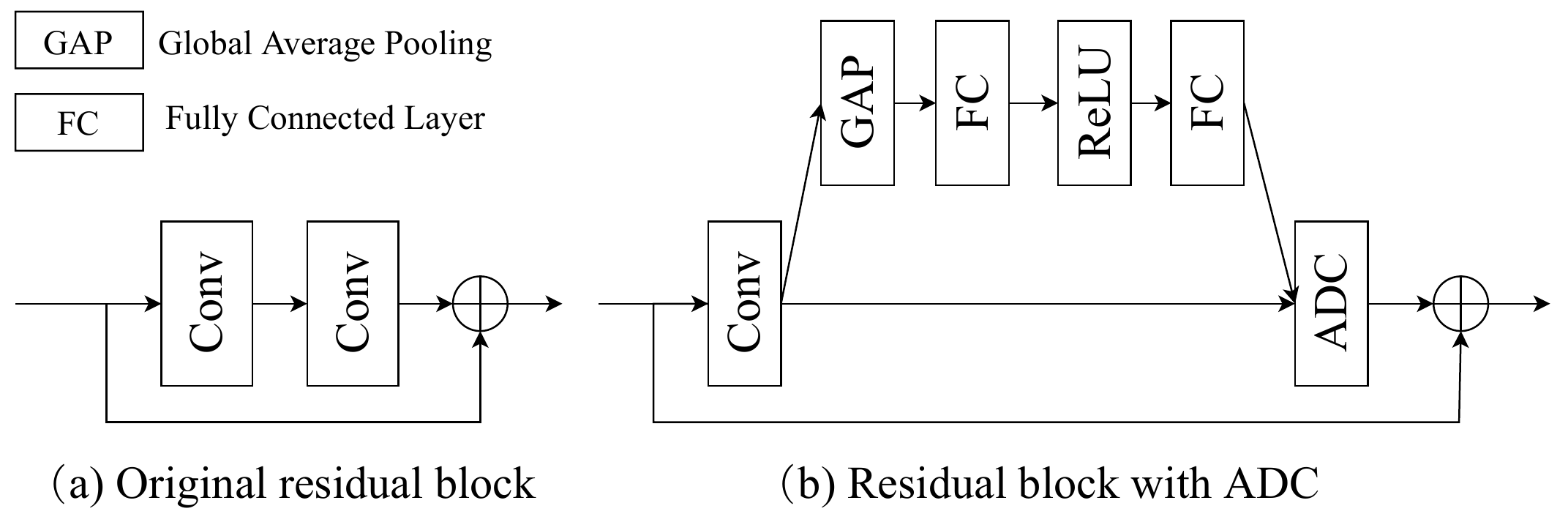}
	\caption{We replace one convolution layer in original residual block (shown in (a)) with ADC (shown in  (b)).}~\label{res_block}
\end{figure}
\subsection{Instantiation}
We plug ADC into the backbones of frequently used HPE models, including the family of SimpleBaseline~\cite{simplebaseline} and HRNet~\cite{hrnet}. Their backbones are built up with residual blocks~\cite{resnet}. As shown in Figure~\ref{res_block}, we replace one ordinary convolution layer in the original residual block by ADC. The weights of the last layer in the dilated-rates regression module are initialized as zeros and its bias are initialized as ones. Thus, the generated dilation rates in ADC are initialized are ones. The dilation groups $g$ are set as $g=C_{in}$, in which case each group contains only one channel. Thus every channel can exploit context information at different scales, and ADC could fuse as much richer multi-scale information as it can. We also experimentally demonstrate that the performance is positively correlated to $g$ in Sec~\ref{sec:aba_g}.

\section{Experiments}
\subsection{Experiments on COCO}
\noindent
\textbf{Dataset.} All of our experiments about human pose estimation are done on COCO dataset~\cite{coco}. It contains over $200K$ persons and $250K$ persons. Our models are trained on COCO train2017 ($57K$ images), and evaluated on COCO val2017 ($5K$ images) and COCO test-dev ($20K$ images). 

\vspace{0.02\linewidth}
\noindent
\textbf{Evaluation metric.} We use the standard evaluation metric Object Keypoint Similarity (OKS) to evaluate our models. $\text{OKS}={\frac{\sum_{i}\exp(-d_i^2/2s^2k_i^2)\delta(v_i>0)}{\sum_{i}\delta(v_i>0)}}$, where $d_i$ is the Euclidean distance between the detected keypoint and its corresponding ground-truth, $v_i$ is the visibility flag of the ground-truth, $s$ denotes the person scale, and $k_i$ is a per-keypoint constant that controls falloff. We report the standard average precision ($AP$) and recall, including $AP^{50}$ ($AP$ at OKS=$0.5$), $AP^{75}$, $AP$ (mean of $AP$ scores from OKS=$0.50$ to OKS=$0.95$ with the increment as $0.05$, $AP^{M}$ ($AP$ scores for person of medium sizes) and $AP^{L}$ ($AP$ scores for persons of large sizes).

\vspace{0.02\linewidth}
\noindent
\textbf{Training.} Following the setting of~\cite{hrnet}, we augment the data by random rotation ([$-30^\circ$, $30^\circ$]), random scaling ([$0.7$, $1.3$]), random translation ([$-40$, $40$]), random horizontal flip and half body transform~\cite{half_body}. Then we crop out each single person according to their ground-truth bounding boxes. These crops are resized to $256\times192$ (or $384\times288$) and input to the HPE model. 

The models are optimized by Adam~\cite{adam} optimizer, and the initial learning rate is set as $1\times10^{-3}$. For the family of HRNet, each model is trained for 210 epochs and the learning rate decays to $1\times10^{-4}$ and $1\times10^{-5}$ at $170th$ and $200th$ epoch respectively. For the family of SimpleBaseline, each model is trained for 140 epochs and the learning rate decays to $1\times10^{-4}$ and $1\times10^{-5}$ at $90th$ and $110th$ epoch respectively. All models are trained and tested on $4$ Tesla V100 GPUs. More details can be referred to the Github repository Pose\footnote{\url{https://github.com/leoxiaobin/deep-high-resolution-net.pytorch.git}}.

\vspace{0.02\linewidth}
\noindent
\textbf{Testing.} During testing, we use the same person detection results provided in~\cite{simplebaseline}, which are widely used for many single-person HPE models~\cite{hrnet,rsn}. Single persons are cropped out according to the detection results and then resized and input to the HPE models. The flip test~\cite{hrnet} is also performed in all experiments. Each keypoint location is predicted by adjusting the highest heatvalue location with a quarter offset in the direction from the highest response to the second-highest response~\cite{hrnet}.

\subsubsection{Ablation Study} To fully demonstrate the superiority of ADC, we perform ablation studies on different models, including the family of SimpleBaseline~\cite{simplebaseline} and HRNet~\cite{hrnet}. The results are shown in Table~\ref{aba_adc}. As one can see, ADC can bring consistent improvement for different models. For the smallest model, \ie SimpleBaseline-Res50, ADC brings an improvement of $1.4$ on $AP$ score. For the largest model, \ie HRNet-W48, there is still an improvement of $0.4$ on $AP$ score. The increments decay as the $AP$ scores increase. This may because it is harder to improve the performance of a more accurate model.  From $AP^{M}$ and $AP^{L}$, we can see that the improvements in medium and large persons are roughly the same. It indicates that ADC benefits equally the keypoint detection of large and medium persons.

\begin{figure}[t]
	\centering
	\includegraphics[width=\linewidth]{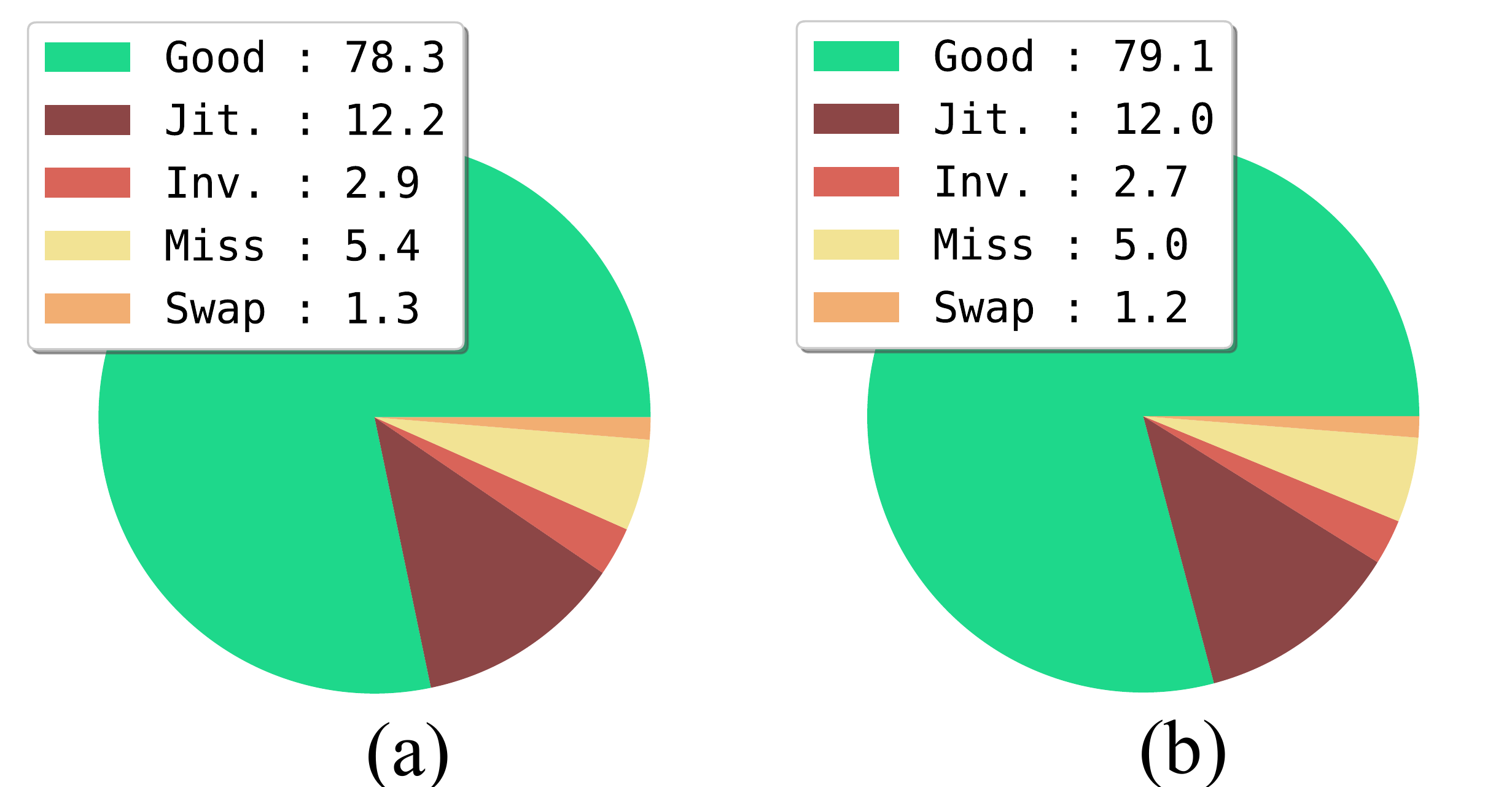}
	\caption{Error analysis results of SimpleBaseline-Res50 (a) without ADC and (b) with ADC.}~\label{error_plot}
	\vspace{-0.05\linewidth}
\end{figure}

\begin{figure*}[t]
	\centering
	\includegraphics[width=\linewidth]{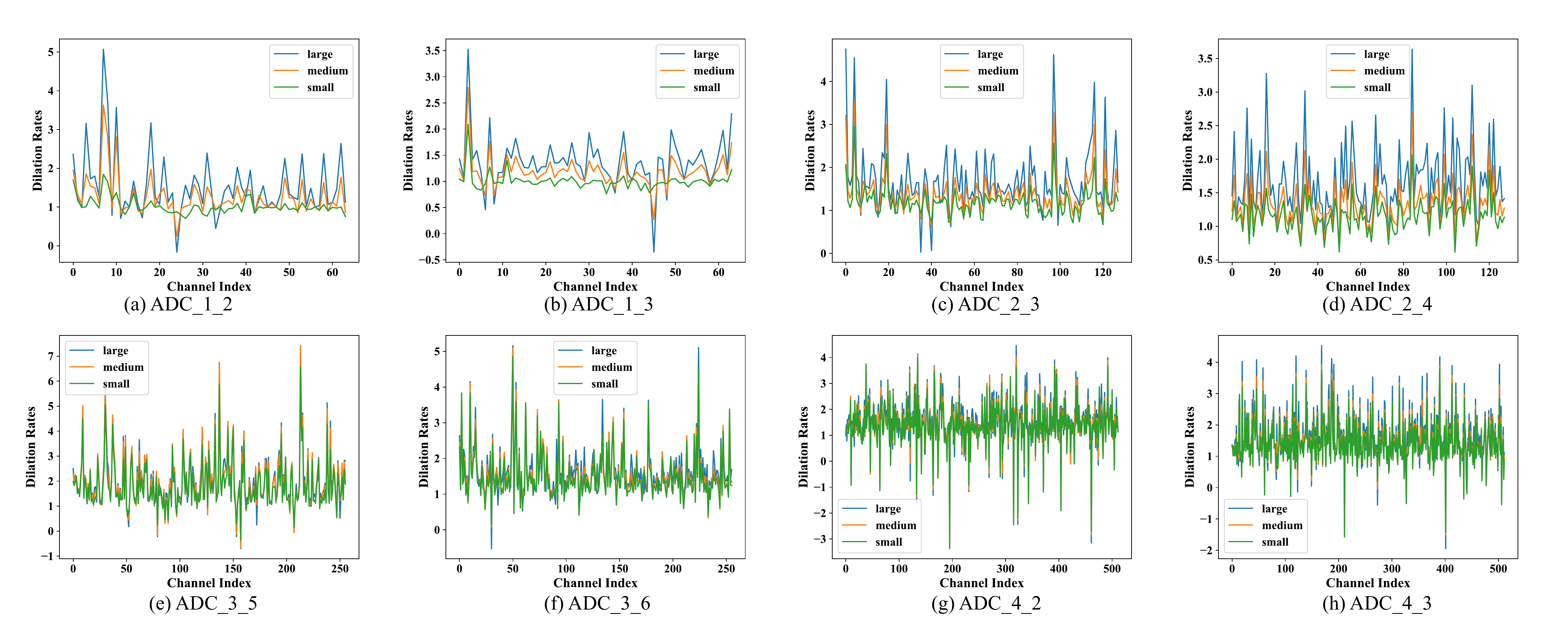}
	\caption{Mean dilation rates in ADC of different blocks in SimpleBaseline-Res50. The subplots are named in the format of ADC\_stageID\_blockID. These statistical comparisons suggest that the dilation rates for relatively larger persons are also likely to be larger.}~\label{rates_stat}
\end{figure*}

\subsubsection{Error Analysis} In this section, we use the error analysis tool in~\cite{coco_ana} to further explore how ADC help HPE models achieve better results. We mainly study four types of errors: 1) \textit{jitter}: small error around the correct keypoint location; 2) \textit{missing}: large localization error, the detected keypoint is not within the proximity of any body part; 3) \textit{inversion}: confusion between semantically similar parts belonging to the same instance. The detection is in the proximity of the true keypoint location of the wrong body part; 4) \textit{swap}: confusion between semantically similar parts of different instances. The detection is within the proximity of a body part belonging to a different person. We use SimpleBaseline-Res50 as the baseline model, and plot the error analysis results with and without ADC in Figure~\ref{error_plot}.  As one can see, ADC can reduce the proportion of all four types of errors. Especially, the proportion of missing error is reduced by $0.4\%$. It suggests that ADC could help the model to be more robust and detect keypoints in more cases. This may be attributed to that ADC can adaptively adjust the dilation rates. The jitter error and inversion error directly indicate the accuracy of location and classification respectively. The proportions of these two errors are both reduced by $0.2\%$. It suggests that ADC can simultaneously benefit the location and classification of keypoints. 

\subsubsection{Statistical Analysis}
In this section, we make a statistical analysis to further investigate how the generated dilation rates in ADC are related to the sizes of test persons. We divide the test persons in COCO val2017 into three types according to the areas of their bounding boxes. Persons whose bounding boxes have areas: 1) smaller than $32\times32$ are divided into the \textit{small group} ($53166$ persons); 2) greater than $32\times32$ but smaller than $96\times96$ are divided into the \textit{medium group} ($25173$ persons); 3) greater than $96\times96$ are divided into the \textit{large group} ($25787$ persons). We still use SimpleBaseline-Res50 with ADC as the studied model.  The backbone, \ie Resnet-50, has four stages, and they contain $3$, $4$, $6$, $3$ residual blocks respectively. 

We plot the means and variations of the dilation rates of different channels in Figure~\ref{rates_stat}. For example, Figure~\ref{rates_stat} (a) shows the mean dilation rates of ADC in the third block of the first stage ($256$ channels). Figure~\ref{rates_stat} (especially the top row) suggests that the dilation rates are closely related to the sizes of test persons. The dilation rates for larger persons are more likely to larger. It enables ADC to be more robust over various human sizes. Besides, the dilation rates for the deeper block also tend to have larger means (bottom row). It may because that deeper blocks are more concerned with semantic features and need larger dilation rates to enlarge the receptive fields. Additionally, the mean dilation rates of different channels in the same layer are quite different. The larger variance of these dilation rates also indicates that ADC can fuse rich multi-scale information via different channels. 

\subsubsection{Study of Dilation Groups $g$} 
We perform comparative experiments to explore the influence of dilation groups $g$. We use the SimpleBaseline-Res50 as the baseline. We gradually improve $g$ from $2$ to $C_{in}$. The results are shown in Table~\ref{aba_g}. As one can see, the model performance becomes better when $g$ increases. It suggests that the number of different dilation rates matters, which also indicates the importance of multi-scale information fusion in HPE.~\label{sec:aba_g}

\begin{table}[h]
	\centering
	\caption{Results of SimpleBaseline-Res50 with different dilation groups $g$. The input sizes are $256\times192$. Results are reported on COCO val2017.}~\label{aba_g}
	\setlength{\tabcolsep}{0.55cm}
	\resizebox{\linewidth}{!}{
		\begin{tabular}{c|cccc}
			\hline
			Groups    & $2$    & $4$    & $8$    & $C_{in}$ \\ \hline
			$AP$      & $71.5$ & $71.5$ & $71.7$ & $71.8$   \\
			$AP^{50}$ & $89.1$ & $89.3$ & $89.1$ & $89.2$   \\
			$AP^{75}$ & $79.4$ & $79.2$ & $79.5$ & $79.7$   \\
			$AP^{M}$  & $68.2$ & $68.3$ & $68.3$ & $68.5$   \\
			$AP^{L}$  & $78.3$ & $78.3$ & $78.5$ & $78.7$   \\ \hline
	\end{tabular}}
\end{table}


\subsubsection{Compared with Other Methods}~\label{exp_deforms}
In this section, we experimentally prove that ADC is more suitable to human pose estimation than deformable convolution (DC)~\cite{dcn} and scale-adaptive convolution (SAC)~\cite{sac}. Comparative experiments are performed on SimpleBaseline-Res50. As shown in Table~\ref{aba_deform}, although DC can bring an improvement on the baseline, its performance is inferior to that of ADC. As we have discussed in Sec~\ref{dis_deform}, the unconstrained and independent offsets of DC may cause the input and output feature maps to lose their spatial correspondence, which potentially hurt the accuracy of location. SAC can alleviate the size inconsistent along spatial dimensions, but involves little multi-scale fusion along the channel dimension, which is more important in HPE. Consequently, the improvement of SAC is lower than both DC and ADC. 

\begin{table}[h]
	\centering
	\caption{Results of SimpleBaseline-Res50 with deformable convolution (DC) or adaptive dilated convolution (ADC). The input sizes are $256\times192$. Results are reported on COCO val2017.}\label{aba_deform}
	\setlength{\tabcolsep}{0.25cm}
	\resizebox{\linewidth}{!}{
		\begin{tabular}{c|ccccc}
			\hline
			Method   & $AP$   & $AP^{50}$ & $AP^{75}$ & $AP^{M}$ & $AP^{L}$ \\ \hline
			Baseline & $70.4$ & $88.6$    & $78.3$    & $67.1$   & $77.2$   \\
			DC~\cite{dcn}      & $71.4$ & $89.2$    & $79.3$    & $67.9$   & $78.3$   \\
			SAC~\cite{sac}      & $71.1$ & $89.1$    & $78.7$    & $67.7$   & $78.1$   \\
			ADC      & $71.8$ & $89.2$    & $79.7$    & $68.5$   & $78.7$   \\ \hline
	\end{tabular}}
\end{table}

\subsection{Experiments for Semantic Segmentation}
Similar to human pose estimation, semantic segmentation also requires rich multi-scale information to make a balance between local and semantic features. Thus the proposed ADC should also benefit the performance of semantic segmentation models. In this section, we plug ADC into different models to demonstrate its benefits on semantic segmentation. 

We use CityScapes~\cite{cityscapes} as our training (2975 images) and validation (500 images) datasets. We use FCN~\cite{fcn}, PSANet~\cite{psanet}, DeepLabV3~\cite{aspp} and DeepLabV3+~\cite{deeplabv3+} as our baseline models. The input sizes are set as $769\times769$. All models are trained for $40K$ iterations. More details can be  referred to the Github repository mmsegmention\footnote{\url{https://github.com/open-mmlab/mmsegmentation.git}}. As shown in Table~\ref{aba_seg}, ADC can bring consistent improvements to different Semantic Scene Parsing models. For FCN, ADC even improves the  mIOU by $4.27$. 
\begin{table}[h]
	\centering
	\caption{Results (mIOU) of different models on CityScapes validation dataset. The input sizes are $769\times769$. All results are trained for $40K$ iterations. w/o ADC: without ADC. w/ ADC: with ADC.}\label{aba_seg}
	\setlength{\tabcolsep}{0.25cm}
	\resizebox{\linewidth}{!}{
		\begin{tabular}{c|c|cc}
			\hline
			Method     & Backbone & w/o ADC & w/ ADC  \\ \hline
			FCN~\cite{fcn}& Res50    & $71.47$ & $75.74$ \\
			\hline
			PSANet~\cite{psanet}&Res50    & $77.99$ & $78.57$ \\
			\hline
			DeepLabV3~\cite{aspp}& Res50    & $78.58$ & $78.70$ \\
			\hline
			DeepLabV3+~\cite{deeplabv3+}& Res50    & $78.97$ & $80.11$ \\ \hline
	\end{tabular}}
\end{table}

\section{Conclusion}
In this paper,  we mainly focus on multi-scale fusion methods in human pose estimation. Existing HPE methods usually fuse feature maps of different spatial sizes to exploit multi-scale information. However, the location information is irreversibly destroyed during the downscaling, and thus the upscaled feature maps may be not well spatially-aligned. This non-alignment potentially hurts the accuracy of keypoint location. Besides, scales of these feature maps are fixed and inflexible, which may restrict its generalization over different human sizes. In this paper, we propose an adaptive dilated convolution (ADC), which exploits multi-scale information by fusing channels with different dilation rates. In this way, each channel in ADC can represent features at a scale, and thus ADC can exploit richer multi-scale information from features of the same spatial sizes. More importantly, the dilation rates for different channels in ADC are adaptively generated, which enables ADC to adjust the scales according to the sizes of test persons. As a result, ADC can help HPE fuse better aligned and more generalized multi-scale features. Extensive experiments on both human pose estimation and semantic segmentation prove that ADC can bring consistent improvements to these methods.

\section*{Acknowledgements}
This work is jointly supported by National Key Research and Development Program of China (2016YFB1001000), Key Research Program of Frontier Sciences, CAS (ZDBS-LY-JSC032), Shandong Provincial Key Research and Development Program (2019JZZY010119), and CAS-AIR.

\bibliographystyle{IEEEtran}
\bibliography{pose}
\end{document}